\documentclass[10pt,twocolumn,letterpaper]{article}

\usepackage{cvpr}
\usepackage{times}
\usepackage{epsfig}
\usepackage{graphicx}
\usepackage{amsmath}
\usepackage{amssymb}

\usepackage{enumitem}
\usepackage{subcaption}
\usepackage{pifont}
\usepackage{blindtext}
\usepackage{morewrites}
\usepackage{color,soul}

\usepackage{array}
\usepackage{algorithm}
\usepackage{algcompatible}
\usepackage{comment}
\newcommand{\xmark}{\ding{55}}%
\newcommand{\cmark}{\ding{51}}%
\usepackage{makecell}
\makeatletter
\renewcommand{\paragraph}{%
  \@startsection{paragraph}{4}%
  {\z@}{0.5ex \@plus 1ex \@minus .2ex}{-1em}%
  {\normalfont\normalsize\bfseries}%
}
\makeatother
\usepackage[breaklinks=true,bookmarks=false]{hyperref}

\cvprfinalcopy

\ifcvprfinal\fi
\begin{document}

\title{Automatic Face Aging in Videos via Deep Reinforcement Learning}

\author{Chi Nhan Duong $^{1}$, Khoa Luu $^{2}$, Kha Gia Quach $^{1}$, Nghia Nguyen $^{2}$, \\Eric Patterson $^{3}$, Tien D. Bui $^{1}$, Ngan Le $^{4}$\\
	$^{1}$ Computer Science and Software Engineering, Concordia University, Canada\\
	$^{2}$ Computer Science and Computer Engineering, University of Arkansas, USA\\
	$^{3}$ School of Computing, Clemson University, USA\\
	$^{4}$ Electrical and Computer Engineering, Carnegie Mellon University, USA\\
	\tt\small $^{1}$\{dcnhan, kquach\}@ieee.org, bui@encs.concordia.ca, $^{2}$\{khoaluu, nhnguyen\}@uark.edu, \\ \tt\small$^{3}$ekp@clemson.edu,   $^{4}$thihoanl@andrew.cmu.edu
}

\maketitle

\begin{abstract}
This paper presents a novel approach to synthesize automatically age-progressed facial images in video sequences using Deep Reinforcement Learning. The proposed method models facial structures and the longitudinal face-aging process of given subjects coherently across video frames. The approach is optimized using a long-term reward,  Reinforcement Learning function with deep feature extraction from Deep Convolutional Neural Network. Unlike previous age-progression methods that are only able to synthesize an aged likeness of a face from a single input image, the proposed approach is capable of age-progressing facial likenesses in videos with consistently synthesized facial features across frames. In addition, the deep reinforcement learning method guarantees preservation of the visual identity of input faces after age-progression. Results on videos of our new collected aging face AGFW-v2 database demonstrate the advantages of the  proposed solution in terms of both quality of age-progressed faces, temporal smoothness, and cross-age face verification.
\end{abstract}

\begin{figure}[t]
	\centering \includegraphics[width=0.96\columnwidth]{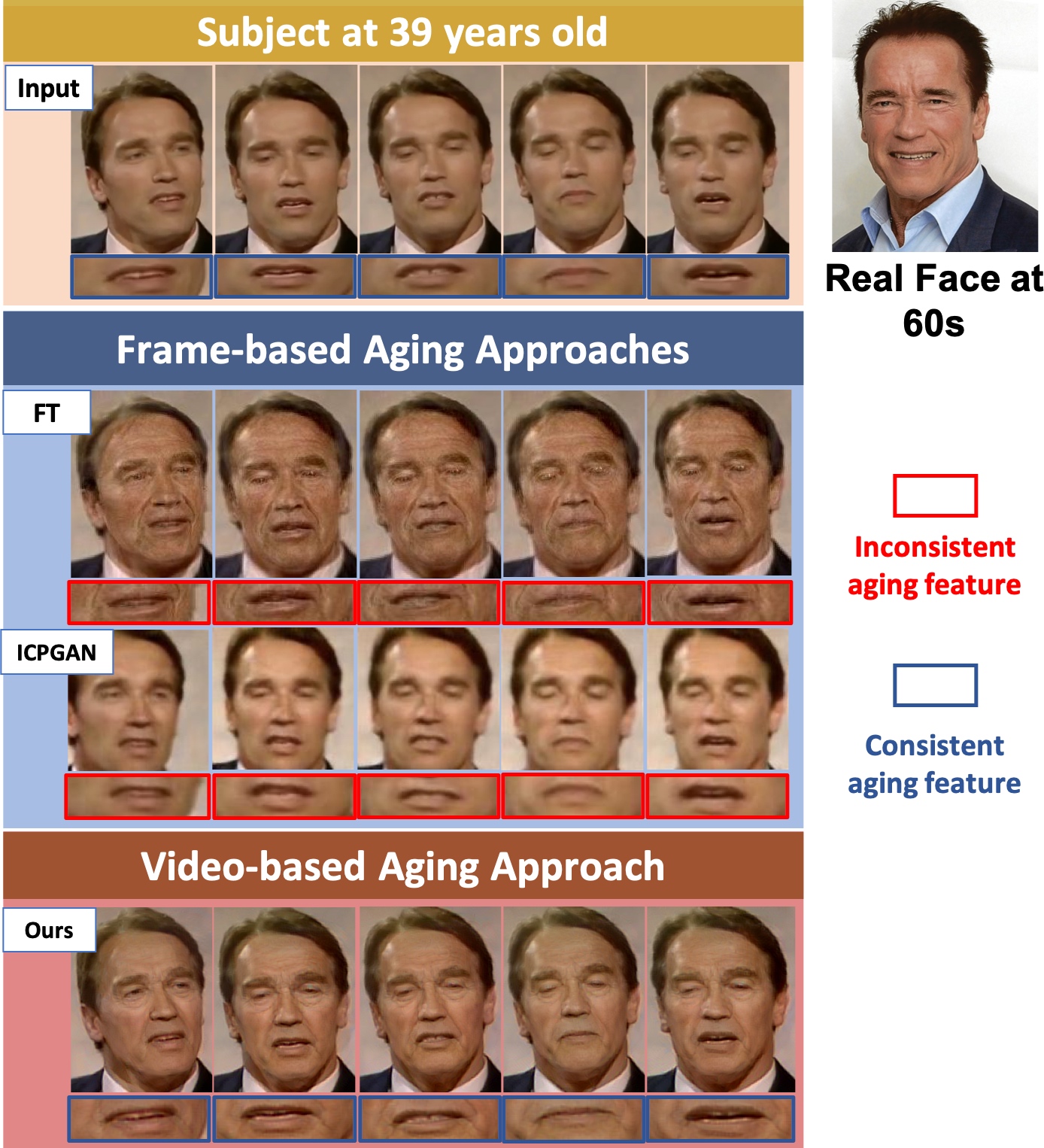}
	\caption{Given an input video, while frame-based approaches produce inconsistent aging features, our video-based method ensures consistency among video frames.}
	\label{fig:FrameVsVideo_Mark}
\end{figure}

\begin{table*} [!t]
	\small
	\centering
	\caption{The comparison of the properties between our video-based approach and other age progression methods.} 
	\begin{tabular}{ >{\arraybackslash}m{3cm} p{1.66cm} p{1.65cm}  p{1.8cm} p{1.65cm} p{1.65cm} p{1.8cm} }
		\Xhline{2\arrayrulewidth}
		& \textbf{Ours} & ICPGAN \cite{wang2018face_aging}& TNVP \cite{Duong_2017_ICCV} & CAAE \cite{Zhang_2017_CVPR}  & RFA \cite{wang2016recurrent} & TRBM \cite{Duong_2016_CVPR} \\
		\Xhline{2\arrayrulewidth}
		\textbf{Modality} & \textbf{Video-based} & Image-based & Image-based & Image-based & Image-based & Image-based \\
		\textbf{Temporal Consistency} & \textbf{Yes} & No & No & No & No & No \\
		\hline
		\textbf{Aging Mechanism} & \textbf{One-shot} & One-shot & Multiple-shot & One-shot & One-shot & Multiple-shot \\
		\textbf{Architecture} &\textbf{DL + RL} &DL &DL
		& DL & DL & DL \\
		\textbf{Tractability} &\cmark &\cmark & \cmark & \cmark & \cmark & \xmark \\
		\Xhline{2\arrayrulewidth}
	\end{tabular}\label{tab:TenMethodSumm}
	\vspace{-4mm}
\end{table*}

\section{Introduction}
Age-related facial technologies generally address the two areas of age estimation \cite{Chen_FG2011, Luu_FG2011, Luu_BTAS2009, Luu_IJCB2011, Duong_ICASSP2011, Luu_ROBUST2008} and age progression \cite{nhan2015beyond, patterson2007comparison, Zhang_2017_CVPR, Patterson2013, wang2018face_aging, Shu_2015_ICCV}. The face age-estimation problem is defined as building computer software that has the ability to recognize the ages of individuals in a given photograph. Comparatively, the face age-progression problem necessitates the more complex capability to predict the future facial likeness of people appearing in images \cite{Luu_CAI2011}. Aside from the innate curiosity of individuals, research of face aging has its origins in cases of missing persons and wanted fugitives, in either case law enforcement desires plausible age-progressed images to facilitate searches. Accurate face aging also provides benefits for numerous practical applications such as age-invariant face recognition \cite{Xu_IJCB2011, Xu_TIP2015, Le_JPR2015}. There have been numerous anthropological, forensic, computer-aided, and computer-automated approaches to facial age-progression. However, the results from previous methods for synthesizing aged faces that represent accurate physical processes involved in human aging are still far from perfect. This is especially so in age-progressing videos of faces, due to the usual challenges for face processing involving pose, illumination, and environment variation as well as differences between video frames.

There have been two key research directions in age progression for both conventional computer-vision approaches and recent deep-learning methods -- \textit{one-shot synthesis} and \textit{multiple-shot synthesis}.  Both approaches have used facial image databases with longitudinal sample photos of individuals, where the techniques attempt to discover aging patterns demonstrated over individuals or the population represented.  In one-shot synthesis approaches, a new face at the target age is directly synthesized via inferring the relationships between training images and their corresponding age labels then applying them to generate the aged likeness. These prototyping methods \cite{burt1995perception, kemelmacher2014illumination,rowland1995manipulating} often classify training images in facial image databases into age groups according to labels. Then the average faces, or mean faces, are computed to represent the key presentation or archetype of their groups.
The variation between the input age and the target age archetypes is complimented to the input image to synthesize the age-progressed faces at the requested age.
In a similar way, Generative Adversarial Networks (GANs) \cite{Zhang_2017_CVPR, wang2018face_aging} methods present the relationship between semantic representation of input faces and age labels by constructing a deep neural network generator. It is then combined with the target age labels to synthesize output results.

Meanwhile, in multiple-shot synthesis, the longitudinal aging process is decomposed into multiple steps of aging effects \cite{Duong_2017_ICCV,Duong_2016_CVPR, Shu_2015_ICCV, wang2016recurrent,yang2016face}. These methods build on the facial aging transformation between two consecutive age groups. Finally, the progressed faces from one age group to the next are synthesized step-by-step until they reach the target age. These methods can model the long-term sequence of face aging using this strategy. However, these methods still have drawbacks due to the limitations of long-term aging not being well represented nor balanced in face databases. 

Existing age-progression methods all similarly suffer from problems in both directions. Firstly, they only work on single input images. Supposing there is a need to synthesize aging faces presented in a captured video, these methods usually have to split the input video into separate frames and synthesize every face in each frame \textit{independently} which may often present \textit{inconsistencies} between  synthesized faces. Since face images for each frame are synthesized separately, the aging patterns of generated faces of the same subject are also likely not coherent. Furthermore, most aging methods are unable to produce \textit{high-resolution} images of age progression, important for features such as fine lines that develop fairly early in the aging process.  This may be especially true in the latent based methods \cite{kemelmacher2014illumination, Duong_2017_ICCV,Duong_2016_CVPR, Shu_2015_ICCV, wang2016recurrent,yang2016face}.

\paragraph{Contributions of this work:} 
This paper presents a deep Reinforcement Learning (RL) approach to Video Age Progression to guarantee the consistency of aging patterns in synthesized faces captured in videos. In this approach, the age-transformation embedding is modeled as the optimal selection using Convolutional Neural Network (CNN) features under a RL framework.  Rather than applying the image-based age progression to each video frame independently as in previous methods, the proposed approach has the capability of exploiting the temporal relationship between two consecutive frames of the video. This property facilitates maintaining consistency of aging information embedded into each frame.
In the proposed structure, not only can a \textit{smoother synthesis} be produced across frames in videos, but also the \textit{visual fidelity} of aging data, i.e. all images of a subject in different or the same age, is preserved for better age transformations. To the best of our knowledge, our framework is one of the first face aging approaches in videos.
Finally, this work contributes a new large-scale face-aging database\footnote{\url{https://face-aging.github.io/RL-VAP/}} to support future studies related to automated face age-progression and age estimation in both images and videos.
\section{Related work}

\begin{table*}[!t] 
	\small 
	\centering
	\caption{The properties of our collected AGFW-v2 in comparison with other aging databases. For AGFW-v2 video set, the images of the subjects in old age are also collected for reference in terms of subject's appearance changing.}
	\label{tab:AgingDatabaseProperties}
	\begin{tabular}{l c c c c c c }
		\Xhline{2\arrayrulewidth}
		\textbf{Database} & \textbf{\# Images} & \textbf{\# Subjects} & \textbf{Label type} &  \textbf{Image type} & \textbf{Subject type} & \textbf{Type}\\  
		\Xhline{2\arrayrulewidth}
		MORPH - Album 1 \cite{ricanek2006morph} & 1,690 & 628 & Years old & Mugshot &  Non-famous  & Image DB\\				
		MORPH - Album 2 \cite{ricanek2006morph} & 55,134 & 13,000 & Years old & Mugshot &  Non-famous  & Image DB\\
		\hline
		FG-NET \cite{fgNetData} & 1,002 & 82 & Years old & In-the-wild &  Non-famous  & Image DB\\
		AdienceFaces \cite{levi2015age} & 26,580 & 2,984 & Age groups & In-the-wild &  Non-famous & Image DB\\
		CACD \cite{chen14cross} & 163,446 & 2,000 & Years old & In-the-wild & Celebrities  & Image DB\\
		IMDB-WIKI \cite{Rothe-IJCV-2016} & 52,3051 & 20,284 & Years old & In-the-wild &  Celebrities  & Image DB\\
        
        AgeDB \cite{AgeDB} & 16,488 & 568 & Years old & In-the-wild & Celebrities  & Image DB\\
        AGFW \cite{Duong_2016_CVPR} & 18,685 & 14,185 & Age groups & In-the-wild/Mugshot & Non-famous  & Image DB\\
        \hline
        \textbf{AGFW-v2 (Image)} & \textbf{36,299} & \textbf{27,688} & \textbf{Age groups} & \textbf{In-the-wild/Mugshot} & \textbf{Non-famous}  & \textbf{Image DB}\\
        \textbf{AGFW-v2 (Video)} & \textbf{20,000} & \textbf{100} & \textbf{Years old} & \textbf{Interview/Movie-style} & \textbf{Celebrities}  & \textbf{Video DB}\\
		\hline
		
	\end{tabular}
	\vspace{-4mm}
\end{table*}

This section provides an overview of recent approaches for age progression; \textit{these methods primarily use still images}. The approaches generally fall into one of four groups, i.e. modeling, reconstruction, prototyping, and deep learning-based approaches.

\textit{Modeling-based} approaches aim at modeling both shape and texture of facial images using parameterization method, then learning to change these parameters via an aging function. 
Active Appearance Models (AAMs) have been used with four aging functions in \cite{lanitis2002toward,patterson2006automatic} to model linearly both the general and the specific aging processes. Familial facial cues were combined with AAM-based techniques in \cite{luu2009Automatic, patterson2007comparison}. \cite{Patterson2013} incorporated an AAM reconstruction method to the synthesis process for a higher photographic fidelity of aging.  An AGing pattErn Subspace (AGES) \cite{geng2007automatic} was proposed to construct a subspace for aging patterns as a chronological sequence of face images. 
In \cite{tsai2014human}, AGES was enhanced with guidance faces consisting the subject's characteristics for more stable results. 
Three-layer And-Or Graph (AOG) \cite{suo2010compositional, suo2012concatenational} was used to model a face as a combination of smaller parts, i.e. eyes, nose, mouth, etc. 
Then a Markov chain was employed to learn the aging process for each part. 

In \textit{reconstruction-based} approaches, an aging basis is unified in each group to model aging faces.  Person-specific and age-specific factors were independently represented by sparse-representation hidden factor analysis (HFA) \cite{yang2016face}. 
Aging dictionaries (CDL) \cite{Shu_2015_ICCV} were proposed to model personalized aging patterns by attempting to preserve distinct facial features of an individual through the aging process.

\textit{Prototyping-based} approaches employed proto-typical facial images in a method to synthesize faces. The average face of each age group is used as the representative image for that group, and these are named the ``age prototypes'' \cite{rowland1995manipulating}. Then, by computing the differences between the prototypes of two age groups, an input face can be progressed to the target age through image-based manipulation \cite{burt1995perception}. In \cite{kemelmacher2014illumination}, high quality average prototypes constructed from a large-scale dataset were employed in conjunction with the subspace alignment and illumination normalization.

Recently, \textit{Deep learning-based approaches} have yielded promising results in facial age progression. 
Temporal and Spatial Restricted Boltzmann Machines (TRBM) were introduced in \cite{Duong_2016_CVPR} to represent the non-linear aging process, with geometry constraints, and to model a sequence of reference faces as well as wrinkles of adult faces. A Recurrent Neural Network (RNN) with two-layer Gated Recurrent Unit (GRU) was employed to approximate aging sequences \cite{wang2016recurrent}. 
Also, the structure of Conditional Adversarial Autoencoder (CAAE) was applied to synthesize aged images in \cite{antipov2017face}. Identity-Preserved Conditional Generative Adversarial Networks (IPCGANs) \cite{wang2018face_aging} brought the structure of Conditional GANs with perceptual loss into place for synthesis process. A novel generative probabilistic model, called Temporal Non-Volume Preserving (TNVP) transformation \cite{Duong_2017_ICCV} was proposed to model a long-term facial aging as a sequence of short-term stages. 

\begin{figure*}[t]
	\centering \includegraphics[width=1.5\columnwidth]{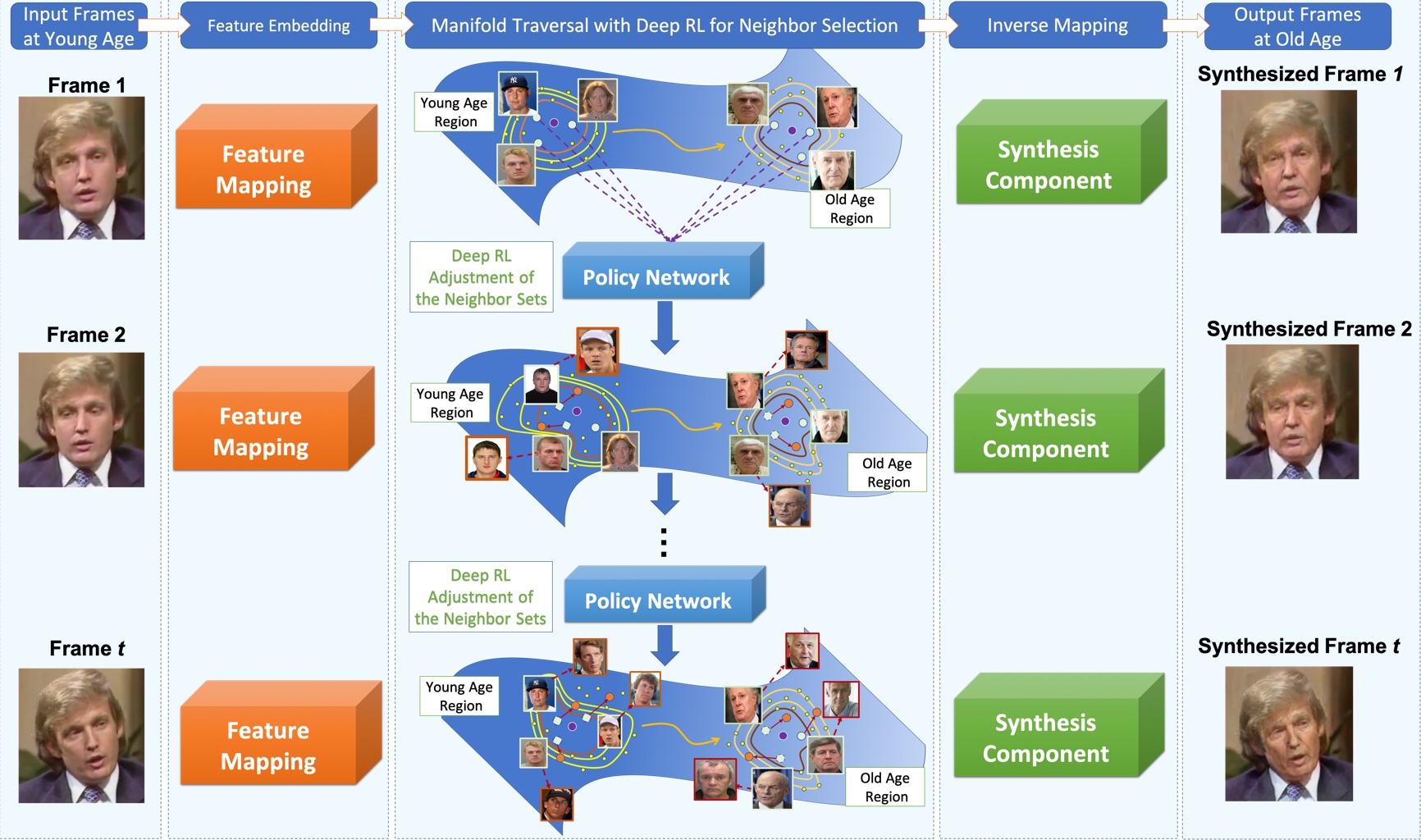}
	\caption{The structure of the face aging framework in video. \textbf{Best viewed in color and 2$\times$ zoom in.}}	
	\label{fig:RL_Framework}
\end{figure*}

\section{Data Collection} \label{sec:dbcollec}
The quality of age representation in a face database is one of the most important features affecting the aging learning process and could include such considerations as the number of longitudinal face-image samples per subject, the number of subjects, the range and distribution of age samples overall, and the population representation presented in the database. 
Previous public databases used for age estimation or progression systems have been very limited in the total number of images, the number of images per subject, or the longitudinal separation of the samples of subjects in the database, i.e. FG-NET \cite{fgNetData}, MORPH \cite{ricanek2006morph}, AgeDB \cite{AgeDB}. Some recent ones may be of larger scale but have noise within the age labels, i.e. CACD \cite{chen14cross}, IMDB-WIKI \cite{Rothe-IJCV-2016}. 
In this work we introduce an extension of Aging Faces in the Wild (AGFW-v2) in terms of both \textit{image and video} collections.
Table \ref{tab:AgingDatabaseProperties} presents the properties of our collected AGFW-v2 in comparison with others.

\subsection{Image dataset}
AGFW \cite{Duong_2016_CVPR} was first introduced with 18,685 images with individual ages sampled ranging from 10 to 64 years old. Based on the collection criteria of AGFW, a double-sized database was desired. Compared to other age-related databases, \textit{most of the subjects in AGFW-v2 are not public figures and less likely to have significant make-up or facial modifications}, helping embed accurate aging effects during the learning process.
In particular, AGFW-v2 is mainly collected from three sources. Firstly, we adopt a search engine using different keywords, e.g. male at 20 years old, etc. Most images come from the daily life of non-famous subjects. Besides the images, all publicly available meta-data related to the subject's age are also collected. 
The second part comes from mugshot images that are accessible from public domain. These are passport-style photos with 
ages reported by service agencies. Finally, we also include the Productive Aging Laboratory (PAL) database \cite{PALDB}.
In total, AGFW-v2 consists of 36,299 images divided into 11 age groups with a span of five years.
\noindent
\subsection{Video dataset}
Along with still photographs, we also collected a video dataset for temporal aging evaluations with 100 videos of celebrities. Each video clip consists of 200 frames.
In particular, searching based on the individuals' names during collection efforts, their interview, presentation, or movie sessions were selected such that only one face, in a clear manner, is presented in the frame.
Age annotations were estimated using the year of the interview session versus the year of birth of the individual. Furthermore, in order to provide a reference for subject's appearance in old age, the face images of these individuals at the current age are also collected and provided as meta-data for the subjects' videos. 

\section{Video-based Facial Aging}

In the simplest approach, age progression of a sequence may be achieved by independently employing image-based aging techniques on each frame of a video. However, treating single frames independently may result in inconsistency of the final aged-progressed likeness in the video, i.e. some synthesized features such as wrinkles appear differently across consecutive video frames as illustrated in Fig. \ref{fig:FrameVsVideo_Mark}. 
Therefore, rather than considering a video as a set of independent frames, this method exploits the temporal relationship between frames of the input video to maintain visually cohesive age information for each frame.
The aging algorithm is formulated as the sequential decision-making process from a goal-oriented agent while interacting with the temporal visual environment. At time sample, the agent integrates related information of the current and previous frames then modifies action accordingly.  The agent receives a scalar reward at each time-step with the goal of maximizing the total long-term aggregate of rewards, emphasizing effective utilization of temporal observations in computing the aging transformation employed on the current frame.

Formally, given an input video, let $\mathcal{I} \in \mathbb{R}^d$ be the image domain and $\mathbf{X}^t = \{\mathbf{x}_y^t,\mathbf{x}_o^t\}$ be an image pair at time-step $t$ consisting of the $t$-th frame $\mathbf{x}_y^t \in \mathcal{I}$ of the video at young age and the synthesized face $\mathbf{x}_o^t \in \mathcal{I}$ at old age.
The goal is to learn a synthesis function $\mathcal{G}$ that maps $\mathbf{x}_y^t$ to $\mathbf{x}_o^t$ as.
\begin{equation}
\footnotesize
\begin{split}
\mathbf{x}_o^t = \mathcal{G}(\mathbf{x}_y^t) | \mathbf{X}^{1:t-1}
\end{split}
\label{eqn:mapping1}
\end{equation}
The conditional term indicates the temporal constraint needs to be considered during the synthesis process. 
To learn $\mathcal{G}$ effectively, we decompose  $\mathcal{G}$ into sub-functions as.
\begin{equation}
\footnotesize
\begin{split}
 \mathcal{G} = \mathcal{F}_1 \circ \mathcal{M} \circ \mathcal{F}_2
 \end{split}    
\end{equation}
where $\mathcal{F}_1: \mathbf{x}_y^t \mapsto \mathcal{F}_1(\mathbf{x}_y^t)$ maps the young face image $\mathbf{x}_y^t$ to its representation in feature domain; $\mathcal{M}: (\mathcal{F}_1(\mathbf{x}_y^t);\mathbf{X}^{1:t-1}) \mapsto \mathcal{F}_1(\mathbf{x}_o^t)$ defines the traversing function in feature domain; and  $\mathcal{F}_2: \mathcal{F}_1(\mathbf{x}_o^t) \mapsto \mathbf{x}_o^t$ is the mapping from feature domain back to image domain.

Based on this decomposition, the architecture of our proposed framework (see Fig. \ref{fig:RL_Framework}) consists of three main processing steps: (1) Feature embedding; (2) Manifold traversal; and (3) Synthesizing final images from updated features.
In the second step, a Deep RL based framework is proposed to guarantee the consistency between video frames in terms of aging changes during synthesis process.

\subsection{Feature Embedding} \label{sec:FeatEmbbed}
The first step of our framework is to learn an embedding function $\mathcal{F}_1$ to map $\mathbf{x}_y^t$ into its latent representation $\mathcal{F}_1(\mathbf{x}_y^t)$. Although there could be various choices for $\mathcal{F}_1$, to produce high quality synthesized images in later steps, the chosen structure for $\mathcal{F}_1$ should produce a feature representation with two main properties: (1) \textit{linearly separable} and (2) \textit{detail preserving}. On one hand, with the former property, transforming the facial likeness from one age group to another age group can be represented as the problem of linearly traversing along the direction of a single vector in feature domain. On the other hand, the latter property guarantees a certain detail to be preserved and produce high quality results. In our framework, CNN structure is used for $\mathcal{F}_1$. 
It is worth noting that there remain some compromises regarding the choice of deep layers used for the representation such that both properties are satisfied. \textit{Linear separability} is preferred in deeper layers further along the linearization process while \textit{details of a face} are usually embedded in more shallow layers \cite{mahendran2015understanding}.
As an effective choice in several image-modification tasks \cite{gatys2015texture, gatys2015neural}, we adopt the normalized VGG-19\footnote{This network is trained on ImageNet for better latent space.} and use the concatenation of three layers  $\{conv3\_1, conv4\_1, conv5\_1\}$ as the feature embedding. 

\begin{figure*}[t]
	\centering 
	\includegraphics[width=1.5\columnwidth]{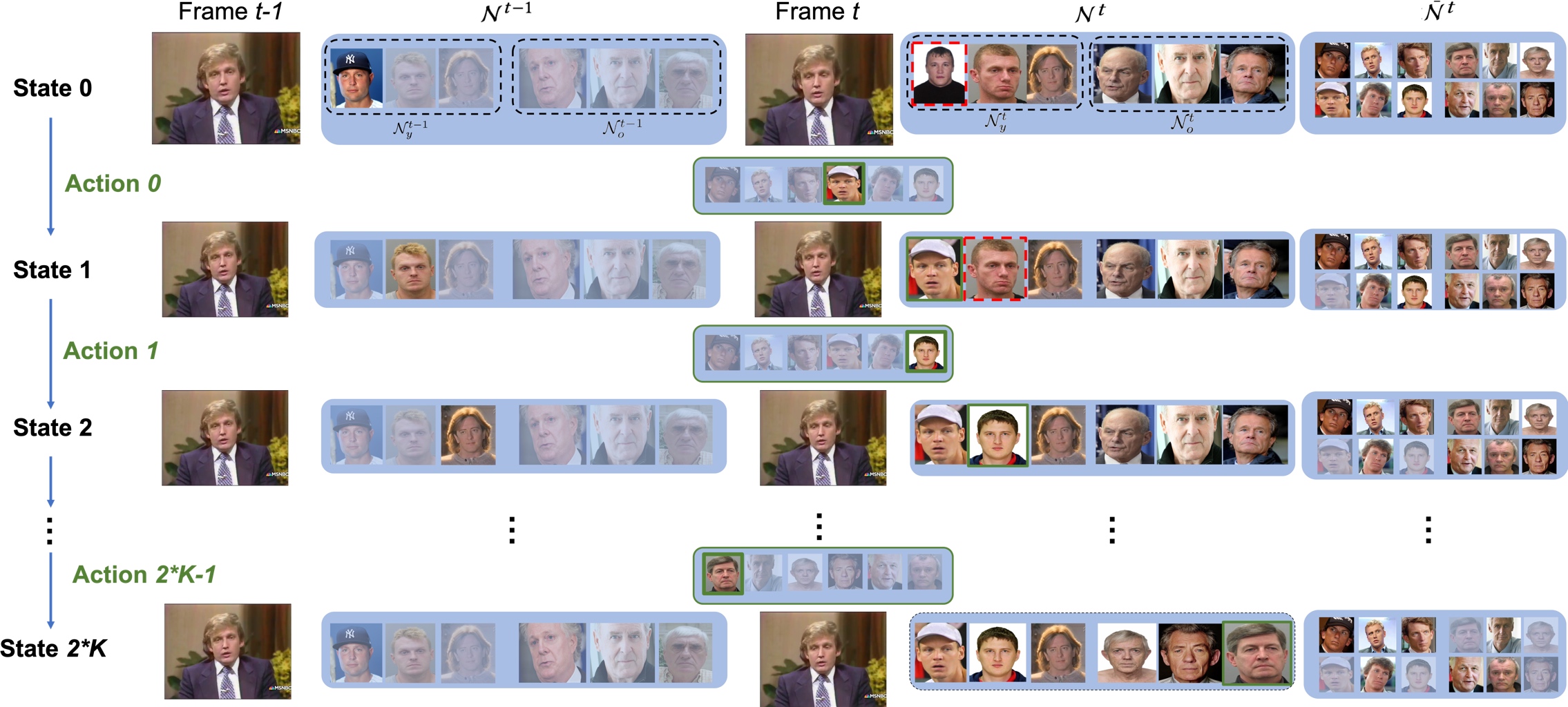}
	\caption{The process of selecting neighbors for age-transformation relationship. \textbf{Best viewed in color and 2$\times$ zoom in.}}	
	\label{fig:RL_policy_net}
\end{figure*}

\subsection{Manifold Traversing}
Given the embedding $\mathcal{F}_1(\mathbf{x}_y^t)$, the age progression process can be interpreted as the linear traversal from the younger age region of $\mathcal{F}_1(\mathbf{x}_y^t)$ toward the older age region of $\mathcal{F}_1(\mathbf{x}_o^t)$ within the deep-feature domain. Then the Manifold Traversing function $\mathcal{M}$ can be written as in Eqn \eqref{eqn:traversing}.
\begin{equation}
\footnotesize
\begin{split}
 \mathcal{F}_1(\mathbf{x}_o^t) & = \mathcal{M}(\mathcal{F}_1(\mathbf{x}_y^t); \mathbf{X}^{1:t-1}) \\
& = \mathcal{F}_1(\mathbf{x}_y^t) + \alpha \Delta^{\mathbf{x}^t|\mathbf{X}^{1:t-1}}
\end{split}
\label{eqn:traversing}
\end{equation}
where $\alpha$ denotes the user-defined combination factor, and $\Delta^{\mathbf{x}^t|\mathbf{X}^{1:t-1}}$ encodes the amount of aging information needed to reach the older age region for the frame $\mathbf{x}_y^t$ conditional on the information of previous frames. 

\subsubsection{Learning from Neighbors} \label{sec:LearnFromNeighbor}
In order to compute $\Delta^{\mathbf{x}^t|\mathbf{X}^{1:t-1}}$ containing only aging effects without the presence of other factors, i.e. identity, pose, etc., we exploit the relationship in terms of the aging changes between the nearest neighbors of $\mathbf{x}_y^t$ in the two age groups. In particular, given $\mathbf{x}_y^t$, we construct two neighbor sets $\mathcal{N}_y^t$ and $\mathcal{N}_o^t$ that contain $K$ nearest neighbors of $\mathbf{x}_y^t$ in the young and old age groups, respectively.
Then 
\small
$\Delta^{\mathbf{x}^t|\mathbf{X}^{1:t-1}}= \Delta^{\mathbf{x}^t|\mathbf{X}^{1:t-1}}_{\mathcal{A}(\cdot, \mathbf{x}_y^t)}$ 
\normalsize
is estimated by:

\begin{equation} \label{eqn:delta_f} \nonumber
\footnotesize
\begin{split}
\Delta^{\mathbf{x}^t|\mathbf{X}^{1:t-1}} 
& =\frac{1}{K} \left[\sum_{\mathbf{x} \in \mathcal{N}_o^t} \mathcal{F}_1(\mathcal{A}(\mathbf{x},\mathbf{x}_y^t))-  \sum_{\mathbf{x} \in \mathcal{N}_y^t} \mathcal{F}_1(\mathcal{A}(\mathbf{x},\mathbf{x}_y^t)) \right]
\end{split}
\end{equation}
\normalsize
where $\mathcal{A}(\mathbf{x},\mathbf{x}_y^t)$ denotes a face-alignment operator that positions the face in $\mathbf{x}$ with respect to the face location in $\mathbf{x}_y^t$. 
Since only the nearest neighbors of $\mathbf{x}_y^t$ are considered in the two sets, conditions apart from age difference should be sufficiently similar between the two sets and subtracted away in $\Delta^{\mathbf{x}^t|\mathbf{X}^{1:t-1}}$. Moreover, the averaging operator also helps to ignore identity-related factors, and, therefore, emphasizing age-related changes as the main source of difference to be encoded in $\Delta^{\mathbf{x}^t|\mathbf{X}^{1:t-1}}$. The remaining question is how to choose the appropriate neighbor sets such that the aging changes provided by $\Delta^{\mathbf{x}^t|\mathbf{X}^{1:t-1}}$ and $\Delta^{\mathbf{x}^{t-1}|\mathbf{X}^{1:t-2}}$ are consistent. In the next section, a Deep RL based framework is proposed for selecting appropriate candidates for these sets.

\subsubsection{Deep RL for Neighbor Selection}
A straightforward technique of choosing the neighbor sets for $\mathbf{x}_y^t$ in young and old age is to select faces that are close to $\mathbf{x}_y^t$ based on some \textit{closeness criteria} such as distance in feature domain, or number of matched attributes. However, since these criteria are not frame-interdependent, they are unable to maintain visually cohesive age information across video frames. 
Therefore, we propose to exploit the relationship presented in the image pair $\{\mathbf{x}_y^t, \mathbf{x}_y^{t-1}\}$ and the neighbor sets of $\mathbf{x}_y^{t-1}$ as an additional guidance for the selection process. Then an RL based framework is proposed and formulated as a sequential decision-making process with the goal of maximizing the temporal reward estimated by the consistency between the neighbor sets of $\mathbf{x}_y^t$ and $\mathbf{x}_y^{t-1}$.

Specifically, given two input frames $\{\mathbf{x}_y^t,\mathbf{x}_y^{t-1}\}$ and two neighbor sets $\{\mathcal{N}_y^{t-1}, \mathcal{N}_o^{t-1}\}$ of $\mathbf{x}_y^{t-1}$, the agent of a policy network will iteratively analyze the role of each neighbor of $\mathbf{x}_y^{t-1}$ in both young and old age in combination with the relationship between $\mathcal{F}_1 (\mathbf{x}_y^t)$ and $\mathcal{F}_1 (\mathbf{x}_y^{t-1})$ to determine new suitable neighbors for $\{\mathcal{N}_y^{t}, \mathcal{N}_o^{t}\}$ of $\mathbf{x}_y^{t}$.
A new neighbor is considered appropriate when it is sufficiently similar to $\mathbf{x}_y^{t}$ and maintains aging consistency between two frames.
Each time a new neighbor is selected, the neighbor sets of $\mathbf{x}_y^t$ are updated and received a reward based on estimating the similarity of embedded aging information between two frames.
As a result, the agent can iteratively explore an optimal route for selecting neighbors to maximize the long-term reward. Fig. \ref{fig:RL_policy_net} illustrates the process of selecting neighbors for age-transformation relationship.

\textbf{State:} The state at $i$-th step $\mathbf{s}^t_i=\left[\mathbf{x}_y^{t}, \mathbf{x}_y^{t-1}, \mathbf{z}^{t-1}_i, (\mathcal{N}^t)_i, \mathcal{\bar{N}}^t, \mathbf{M}_i\right]$ is defined as a composition of six components: (1) the \textit{current frame} $\mathbf{x}_y^t$; (2) the \textit{previous frame} $\mathbf{x}_y^{t-1}$; (3) the \textit{current considered neighbor} $\mathbf{z}^{t-1}_i$ of $\mathbf{x}_y^{t-1}$, i.e. either in young and old age groups; (4) the \textit{current construction of the two neighbor sets} $(\mathcal{N}^t)_i = \{(\mathcal{N}_y^t)_i,(\mathcal{N}_o^t)_i\}$ of $\mathbf{x}_y^{t}$ until step $i$; (5) the \textit{extended neighbor sets} $\mathcal{\bar{N}}^t=\{\mathcal{\bar{N}}_y^t,\mathcal{\bar{N}}_o^t\}$ consisting of $N$ neighbors, i.e. $N > K$, of $\mathbf{x}_y^{t}$ for each age group.
and (6) a \textit{binary mask} $\mathbf{M}_i$ indicating which samples in $\mathcal{\bar{N}}^t$ are already chosen in previous steps. 
Notice that in the initial state  $\mathbf{s}^t_0$, the two neighbor sets  $\{(\mathcal{N}_y^t)_0, (\mathcal{N}_o^t)_0\}$ are initialized using the $K$ nearest neighbors of $\mathbf{x}_y^t$ of the two age groups, respectively. 
Two measurement criteria are considered for finding the nearest neighbors: \textit{the number of matched facial attributes}, e.g gender, expressions, etc.; and \textit{the cosine distance between two feature embedding vectors}.
All values of the mask $\mathbf{M}_i$ are set to 1 in $\mathbf{s}^t_0$.

\textbf{Action:} Using the information from the chosen neighbor $\mathbf{z}^{t-1}_i$ of $\mathbf{x}_y^{t-1}$, and the relationship of $\{\mathbf{x}_y^{t}, \mathbf{x}_y^{t-1}\}$, an action $a_{i}^{t}$ is defined as selecting the new neighbor for the current frame such that with this new sample added to the neighbor sets of the current frame, the aging-synthesis features between $\mathbf{x}_y^{t}$ and $\mathbf{x}_y^{t-1}$ are more consistent. Notice that since not all samples in the database are sufficiently similar to $\mathbf{x}_y^{t}$, we restrict the action space by selecting among $N$ nearest neighbors of $\mathbf{x}_y^{t}$. In our configuration, $N= n * K$ where $n$ and $K$ are set to 4 and 100, respectively.

\textbf{Policy Network:} 
At each time step $i$, the policy network first encodes the information provided in state $\mathbf{s}^t_i$ as
\begin{equation}
\footnotesize
\begin{split}
\mathbf{u}^t_i &= \left[\delta^{\text{pool5}}_{\mathcal{F}_1}(\mathbf{x}_y^t, \mathbf{x}_y^{t-1}), \mathcal{F}^{\text{pool5}}_1(\mathbf{z}_i^{t-1})\right] \\
\mathbf{v}^t_i &=\left[d\left((\mathcal{N}^t)_i,\mathbf{x}_y^t\right), d\left(\mathcal{\bar{N}}^t,\mathbf{x}_y^t\right), \mathbf{M}_i\right]
\end{split}
\end{equation}
where $\mathcal{F}^{\text{pool5}}_1$ is the embedding function as presented in Sec. \ref{sec:FeatEmbbed}, but the $pool5$ layer is used as the representation; $\delta^{\text{pool5}}_{\mathcal{F}_1}(\mathbf{x}_y^t, \mathbf{x}_y^{t-1}) = \mathcal{F}^{\text{pool5}}_1(\mathbf{x}_y^t)-\mathcal{F}^{\text{pool5}}_1(\mathbf{x}_y^{t-1})$ embeds the relationship of $\mathbf{x}_y^{t}$ and $\mathbf{x}_y^{t-1}$ in the feature domain. $d\left((\mathcal{N}^t)_i,\mathbf{x}_y^t\right)$ is the operator that maps all samples in $(\mathcal{N}^t)_i$ to their representation in the form of cosine distance to $\mathbf{x}_y^t$.
The last layer of the policy network is reformulated as $P(\mathbf{z}^t_i = \mathbf{x}_j|\mathbf{s}_{i}^t) = e^{c_i^j} / {\sum_{k} c_i^k}$,
where 
\small
$\mathbf{c}_i = \mathbf{M}_i \odot \left(\mathbf{W} \mathbf{h}_i^t + \mathbf{b}\right)$
\normalsize
and 
\small
$\mathbf{h}_i^t=\mathcal{F}_{\pi}\left(\mathbf{u}_i^t,\mathbf{v}^t_i. \theta_{\pi}\right)$
\normalsize
; $\{\mathbf{W},\mathbf{b}\}$ are weight and bias of the hidden-to-output connections.
Since $\mathbf{h}_i^t$ consists of the features of the sample picked for neighbors of $\mathbf{x}_y^{t-1}$ and the temporal relationship between $\mathbf{x}_y^{t-1}$ and $\mathbf{x}_y^{t}$, it directly encodes the information of \textit{how the face changes} and \textit{what aging information} from the previous frame has been used.
This process helps the agent evaluate its choice to confirm the optimal candidate of $\mathbf{x}_y^t$
to construct the neighbor sets.

The output of the policy network is an $N+1$-dimension vector $\mathbf{p}$ indicating the probabilities of all available actions $P(\mathbf{z}^t_{i}=\mathbf{x}_j|\mathbf{s}_{i}^t),j=1..N$ where each entry indicates the probability of selecting sample $\mathbf{x}_j$ for step $i$. It is noticed that the $N+1$-th value of $\mathbf{p}$ indicates an action that there is no need to update the neighbor sets in this step.
During training, an action $a_{i}^{t}$ is taken by stochastically sampling from this probability distribution. During testing, the one with highest probability is chosen for synthesizing process.

\textbf{State transition:} After decision of action $a_{i}^t$ in state $\mathbf{s}_{i}^t$ has been made, the next state $\mathbf{s}_{i+1}^t$ can be obtained via the state-transition function $\mathbf{s}^t_{i+1} = Transition(\mathbf{s}^t_{i}, a_i^t)$ where $\mathbf{z}^{t-1}_i$ is updated to the next unconsidered sample $\mathbf{z}^{t-1}_{i+1}$ in neighbor sets of $\mathbf{x}_y^{t-1}$. Then the neighbor that is least similar to $\mathbf{x}_y^{t}$ in the corresponding sets of $\mathbf{z}^{t-1}_i$ is replaced by   $\mathbf{x}_j$ according to the action $a_{i}^t$.
The \textit{terminate state} is reached when all the samples of  $\mathcal{N}_y^{t-1}, \mathcal{N}_o^{t-1}$ are considered.

\textbf{Reward:}
During training, the agent will receive a reward signal $r^t_i$ from the environment after executing an action $a_{i}^t$ at step $i$. In our proposed framework, the reward is chosen to measure aging consistency between video frames as.

\begin{equation} \label{eqn:reward}
\footnotesize
r^t_i = \frac{1}{\parallel 
	\Delta^{\mathbf{x}^t|\mathbf{X}^{1:t-1}}_{i,\mathcal{A}(\cdot, \mathbf{x}_y^t)} - \Delta^{\mathbf{x}^{t-1}|\mathbf{X}^{1:t-2}}_{\mathcal{A}(\cdot, \mathbf{x}_y^t)} \parallel + \epsilon}
\end{equation}
Notice that in this formulation, we align all neighbors of both previous and current frames to $\mathbf{x}^t_y$. Since the same alignment operator $\mathcal{A}(\cdot,\mathbf{x}_y^{t})$ on all neighbor sets of both previous and current frames is used, the effect of alignment factors, i.e. poses, expressions, location of the faces, etc., can be minimized in $r^t_i$. Therefore, $r^t_i$ reflects only the difference in aging information embedded into $\mathbf{x}_y^{t}$ and $\mathbf{x}_y^{t-1}$.

\textbf{Model Learning:} The training objective is to maximize the sum of the reward signals: $R = \sum_i r^t_i$. We optimize the recurrent policy network with the REINFORCE algorithm \cite{Williams92simplestatistical} guided by the reward given at each time step. 

\subsection{Synthesizing from Features}
After the neighbor sets of $\mathbf{x}_y^t$ are selected, the $\Delta^{\mathbf{x}^t|\mathbf{X}^{1:t-1}}$ can be computed as presented in Sec. \ref{sec:LearnFromNeighbor} and the embedding of $\mathbf{x}_y^t$ in old age region $\mathcal{F}_1(\mathbf{x}_o^t)$ is estimated via Eqn. \eqref{eqn:traversing}. In the final stage, $\mathcal{F}_1(\mathbf{x}_o^t)$ can then be mapped back into the image domain $\mathcal{I}$ via $\mathcal{F}_2$ which can be achieved by the optimization shown in Eqn. \eqref{eqn:tv_update} \cite{mahendran2015understanding}.
\begin{equation} \label{eqn:tv_update}
\small
\mathbf{x}^{t*}_o = \arg \min_{\mathbf{x}} \frac{1}{2} \parallel \mathcal{F}_1(\mathbf{x}_o^t) - \mathcal{F}_1(\mathbf{x}) \parallel^2_2 + \lambda_{V^\beta} R_{V^\beta}(\mathbf{x})
\end{equation}
where $R_{V^\beta}$ represents the Total Variation regularizer encouraging smooth transitions between pixel values.

\begin{figure}[t]
	\centering \includegraphics[width=0.9\columnwidth]{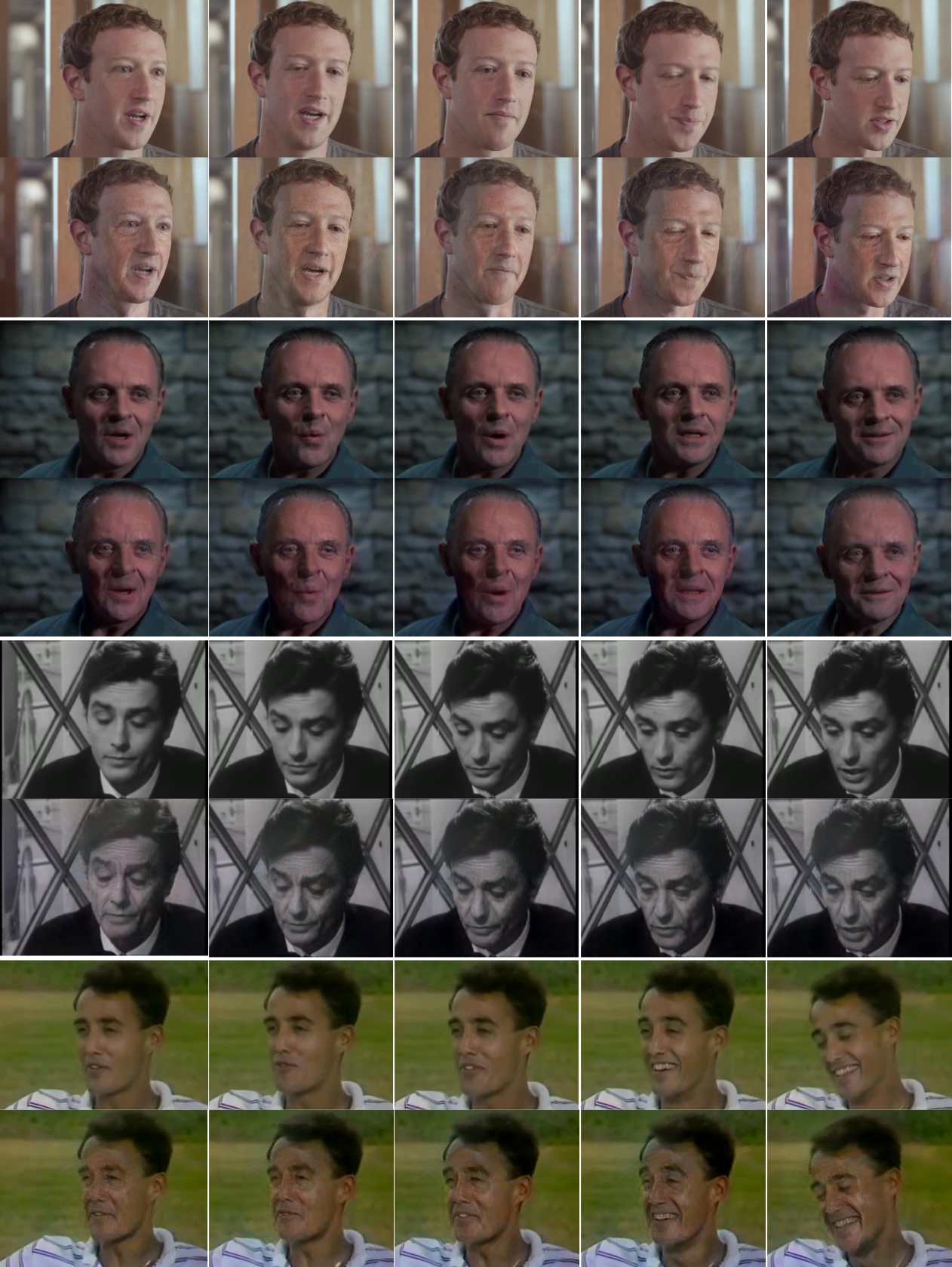}
	\caption{\textbf{Age Progression Results.} For each subject, the two rows shows the input frames at the young age, and the age-progressed faces at 60-years old, respectively.} 
	\label{fig:Video_AP_frontal}
\end{figure}

\section{Experimental Results}

\subsection{Databases} \label{subsec:db}
The proposed approach is trained and evaluated using training and testing databases that are not overlapped. Particularly, the neighbor sets are constructed using a large-scale database composing face images from our collected \textbf{AGFW-v2}
and \textbf{LFW-GOOGLE} \cite{upchurch2016deep}.
Then Policy network is trained using videos from \textbf{300-VW} \cite{shen2015first}. 
Finally, the video set from AGFW-v2 is used for evaluation.

\textbf{LFW-GOOGLE} \cite{upchurch2016deep}: includes 44,697 high resolution images collected using the names of 5,512 celebrities. 
This database does not have age annotation. 
To obtain the age label, we employ the age estimator in \cite{Rothe-IJCV-2016} for initial labels which are manually corrected as needed after estimation. 

\textbf{300-VW} \cite{shen2015first}: includes 218595 frames from 114 videos. Similar to the video set of AGFW-v2, the videos are movie or presentation sessions containing one face per frame.

\begin{figure}[t]
	\centering \includegraphics[width=1\columnwidth]{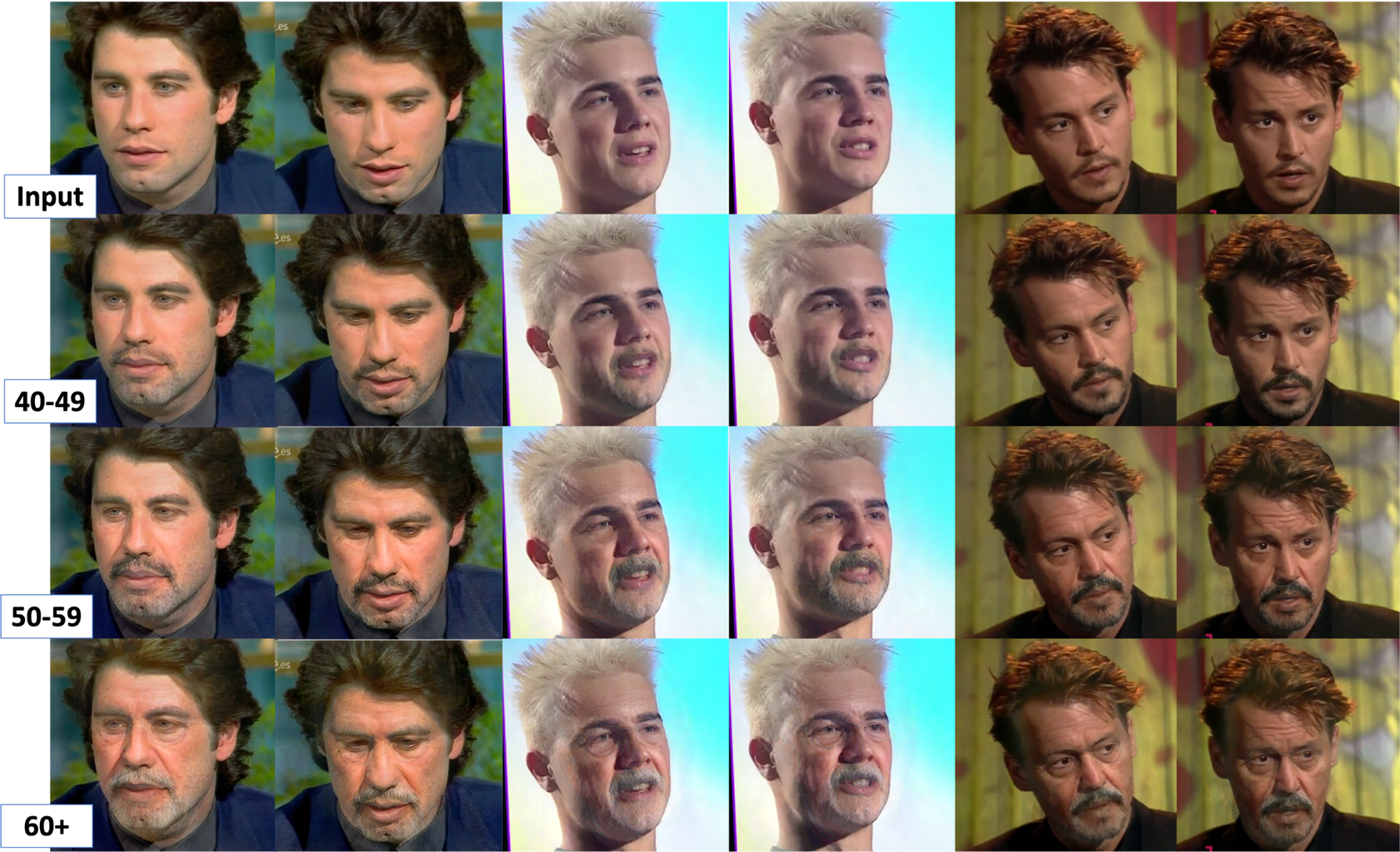}
	\caption{\textbf{Age Progression Results.} Given different frames of a subject, our approach can consistently synthesized the faces of that subject at different age groups.} 	\label{fig:Video_AP_DifferentAgeGroup}
\end{figure}

\subsection{Implementation Details} \label{subsec:imple}

\textbf{Data Setting.} In order to construct the neighbor sets for an input frames in young and old ages, images from AGFW-v2 and LFW-GOOGLE are combined and divided into 11 age groups from 10 to 65 with the age span of five years. 

\textbf{Model Structure and Training.} For the policy network, we employ a neural network with two hidden layers of 4096 and 2048 hidden units, respectively. Rectified Linear Unit (ReLU) activation is adopted for each hidden layer. 
The videos from 300-VW are used to train the policy network.

\textbf{Computational time.} Processing time of the synthesized process depends on the resolution of the input video frames.
It roughly takes from 40 seconds per $240 \times 240$ frame or 4.5 minutes per video frame with the resolution of $900 \times 700$.
We evaluate on a system using an Intel i7-6700 CPU@3.4GHz with an NVIDIA GeForce TITAN X GPU. 

\subsection{Age Progression} \label{subsec:agingresult}
This section demonstrates the validity of the approach for robustly and consistently synthesizing age-progressed faces across consecutive frames of input videos.

\begin{figure}[t]
	\centering \includegraphics[width=1\columnwidth]{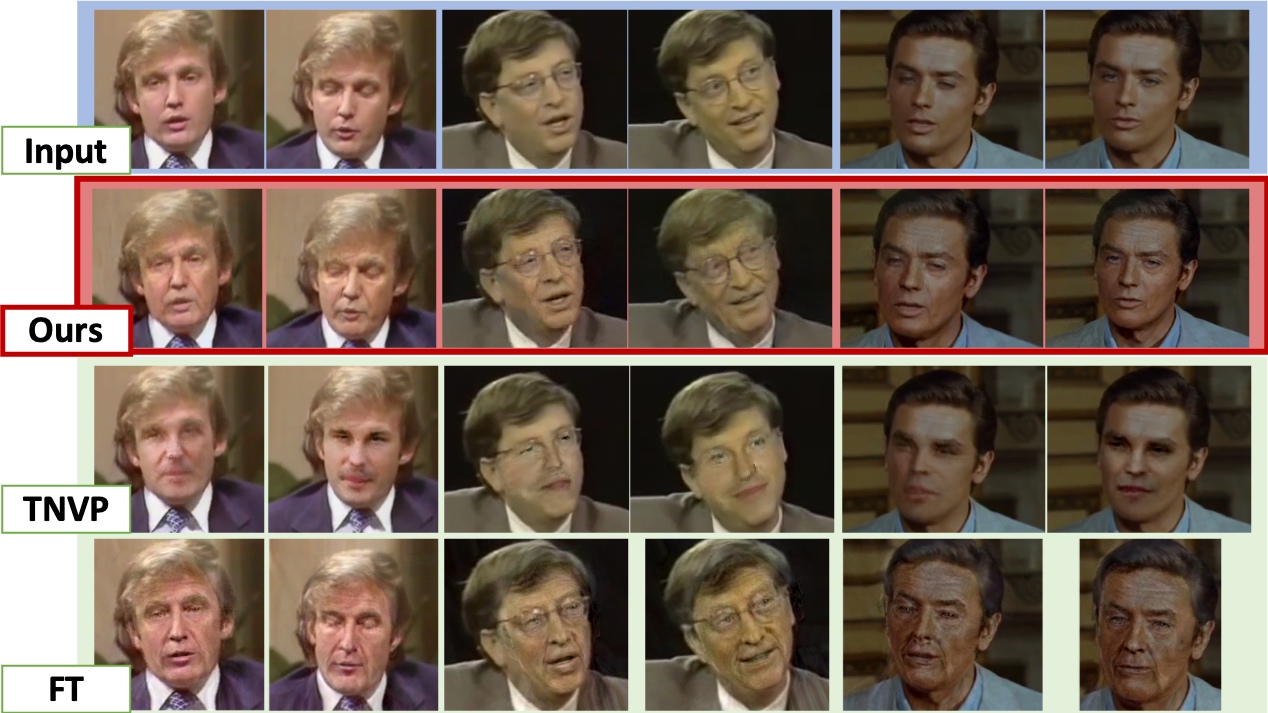}
	\caption{\textbf{Comparisons between age progression approaches}. For each subject, the top row shows frames in the video at a younger age. The next three rows are our results, TNVP \cite{Duong_2017_ICCV} and Face Transformer \cite{faceTransformer}, respectively.}
	\label{fig:AP_Comparisons}
\end{figure}

\textbf{Age Progression in frontal and off-angle faces.} 
Figs. \ref{fig:Video_AP_frontal} and \ref{fig:Video_AP_DifferentAgeGroup} illustrate our age-progression results across frames from AGFW-v2 videos that contain both frontal and off-angle faces. From these results, one can see that even in case of \textit{frontal faces} (i.e. the major changes between frames come from facial expressions and movements of the mouth and lips), or \textit{off-angle faces} (i.e. more challenging due to the pose effects in the combination of other variations), 
our proposed method is able to robustly synthesize aging faces. Wrinkles of soft-tissue areas (i.e. under the subject's eyes; around the cheeks and mouth) are coherent robust between consecutive synthesized frames. We also compare our methods against Temporal Non-volume Preserving (TNVP) approach \cite{Duong_2017_ICCV} and Face Transformer (FT) \cite{faceTransformer} in Fig.  \ref{fig:AP_Comparisons}. These results further show the advantages of our model when both TNVP and FT are unable to ensure the consistencies between frames, and may result in different age-progressed face for each input. Meanwhile, in our results, the temporal information is efficiently exploited. This emphasizes the crucial role of the learned policy network.

\textbf{Aging consistency.} Table \ref{tb:Consistency_Eval} compares the aging consistency between different approaches.
For the \textbf{\textit{consistency measurement}}, we adopt the average inverted reward $r^{-1}$ of all frames for each synthesis video. 
Furthermore, to validate the \textbf{\textit{temporal smoothness}}, we firstly compute the optical flow, i.e. an estimation of image displacements, between frames of each video to estimate changes in pixels through time. Then we evaluate the differences ($\ell_2$-norm) between the flows of the original versus synthesis videos. 
From these results, one can see that  policy network has consistently and robustly shown its role on maintaining an appropriate aging amount embedded to each frame, and, therefore, producing smoother synthesis across frames in the output videos.

\subsection{Video Age-Invariant Face Recognition} \label{subsec:recog}
\begin{figure}[t]
	\centering \includegraphics[width=0.85\columnwidth]{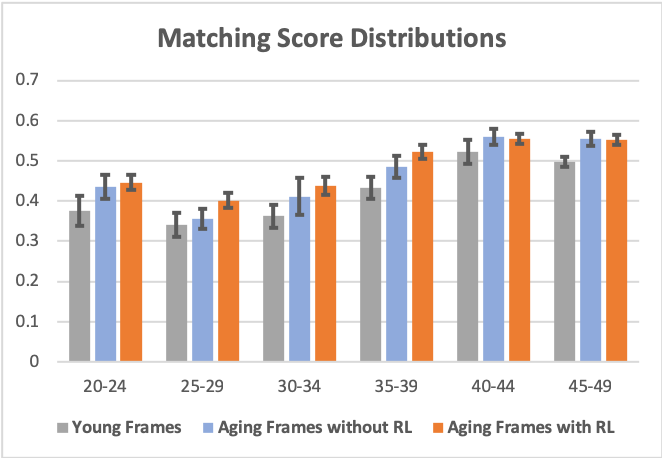}
	\caption{The distributions of the matching scores (of each age group) between frames of original and age-progressed videos against real faces of the subjects at the current age. }	
	\label{fig:MatchingScoreDistribution}
\end{figure}

The effectiveness of our proposed approach is also validated in terms the performance gain for cross-age face verification. With the present of RL approach, not only is the consistency guaranteed, but also are the improvements made in both matching accuracy and matching score deviation. We adapt one of the state-of-the-art deep face matching models in \cite{deng2018arcface} for this experiment. 
We set up the face verification as follows. For all videos with the subject's age labels in the video set of AGFW-v2, the proposed approach is employed to synthesize all video frames to the current ages of the corresponding subjects in the videos. Then each frame of the age-progressed videos is matched against the real face images of the subjects at the current age. The matching scores distributions between original (young) and aged frames are presented in Fig.  \ref{fig:MatchingScoreDistribution}. Compared to the original frames, our age-progressed faces produce higher matching scores and, therefore, improve the matching performance over original frames. Moreover, with the consistency during aging process, the score deviation is maintained to be low. This also helps to improve the overall performance further. The matching accuracy among different approaches is also compared in Table \ref{tb:Consistency_Eval} to emphasize the advantages of our proposed model.

\begin{table}[t]
	\footnotesize
	\centering
	\caption{Comparison results in terms of consistency and temporal smoothness (\textit{smaller value indicates better consistency}); and matching accuracy (\textit{higher value is better}).  
	} 
	\label{tb:Consistency_Eval} 
	\small
	\begin{tabular}{l c c c }
		\Xhline{2\arrayrulewidth}
		\textbf{Method} & 
		\begin{tabular}{@{}c@{}}\textbf{Aging} \\ \textbf{Consistency} \end{tabular}& \begin{tabular}{@{}c@{}}\textbf{Temporal}\\ \textbf{Smoothness}\end{tabular} & \begin{tabular}{@{}c@{}}\textbf{Matching}\\ \textbf{Accuracy}\end{tabular}\\
		\hline
		\begin{tabular}{@{}l@{}} Original Frames\end{tabular} & $-$ & $-$ & 60.61\%\\
		\hline
		FT \cite{faceTransformer} & 378.88 & 85.26 & 67.5\%\\
		TNVP \cite{Duong_2017_ICCV} & 409.45 & 87.01 & 71.57\%\\
		IPCGANs \cite{wang2018face_aging} & 355.91 & 81.45&73.17\%\\
		\hline
		\hline
		\textbf{Ours(Without RL)} & 346.25 & 75.7 & 78.06\%\\
		\textbf{Ours(With RL)} & \textbf{245.64} & \textbf{61.80} & \textbf{83.67\%}\\
		\Xhline{2\arrayrulewidth}
	\end{tabular}
\end{table}

\section{Conclusions}
This work has presented a novel Deep RL based approach for age progression
in videos.
The model inherits the strengths of both recent advances of deep networks and reinforcement learning techniques to synthesize aged faces of given subjects both plausibly and coherently across video frames.
Our method can generate age-progressed facial likenesses in videos with consistently aging features across frames. Moreover, our method guarantees preservation of the subject's visual identity after synthesized aging effects.

{\small
\bibliographystyle{ieee_fullname}
\bibliography{egbib}
}

\end{document}